\documentclass[10pt,twocolumn,letterpaper]{article}

\usepackage{iccv}
\usepackage{times}
\usepackage{epsfig}
\usepackage{graphicx}
\usepackage{amsmath}
\usepackage{amssymb}
\usepackage{tikz}
\usepackage{pgfplots}
\usepackage[normalem]{ulem}
\usepackage{multicol}
\usepackage{booktabs}
\usepackage{caption}
\usetikzlibrary{plotmarks,patterns}
\pgfplotsset{compat=newest}

\usepackage[pagebackref=true,breaklinks=true,letterpaper=true,colorlinks,bookmarks=false]{hyperref}

\iccvfinalcopy 

\newcommand{\figdir}{./figure/}
\definecolor{mygreen}{RGB}{0,135,54}
\definecolor{myblue}{RGB}{33,113,181}
\definecolor{myred}{RGB}{201,0,31}
\definecolor{mypurple}{RGB}{92,58,150}

\ificcvfinal\fi

\begin{document}
	
	\title{Bayesian Loss for Crowd Count Estimation with Point Supervision}
	
	\author{Zhiheng Ma$^1$\footnotemark[1]
		\quad
		Xing Wei$^1$\footnotemark[1]
		\quad
		Xiaopeng Hong$^{1,2}$\footnotemark[2]
		\quad
		Yihong Gong$^1$
		\\
		$^1${Faculty of Electronic and Information Engineering, Xi'an Jiaotong University}
		\\
		$^2${Research Center for Artificial Intelligence, Peng Cheng Laborotory}
		\\ 
		{\tt\small mazhiheng@stu.xjtu.edu.cn}
		\quad
		{\tt\small xingxjtu@gmail.com}
		\quad
		{\tt\small \{hongxiaopeng,ygong\}@mail.xjtu.edu.cn}
	}

	\maketitle
	\ificcvfinal\thispagestyle{empty}\fi
	
	\renewcommand{\thefootnote}{\fnsymbol{footnote}}
	\footnotetext[1]{Equal contribution.} 
	\footnotetext[2]{Corresponding author.}
	
	\begin{abstract}
		In crowd counting datasets, each person is annotated by a point, which is usually the center of the head. And the task is to estimate the total count in a crowd scene. 
		Most of the state-of-the-art methods are based on density map estimation, which convert the sparse point annotations into a ``ground truth" density map through a Gaussian kernel, and then use it as the learning target to train a density map estimator.
		However, such a ``ground-truth" density map is imperfect due to occlusions, perspective effects, variations in object shapes, etc.
		On the contrary, we propose \emph{Bayesian loss}, a novel loss function which constructs a density contribution probability model from the point annotations. Instead of constraining the value at every pixel in the density map, the proposed training loss adopts a more reliable supervision on the count expectation at each annotated point.
		Without bells and whistles, the loss function makes substantial improvements over the baseline loss on all tested datasets. Moreover, our proposed loss function equipped with a standard backbone network, without using any external detectors or multi-scale architectures, plays favourably against the state of the arts. Our method outperforms previous best approaches by a large margin on the latest and largest UCF-QNRF dataset. The source code is available at \url{https://github.com/ZhihengCV/Baysian-Crowd-Counting}.
	\end{abstract}
	
	\section{Introduction}
	
	Counting dense crowds using computer vision techniques has attracted remarkable attentions in recent years. It has a wide range of applications such as estimating the scale of, and counting the number of participants in political rallies, civil unrest, social and sport events, etc.
	In addition, methods for crowd counting also have great potentials to handle similar tasks in other domains, including estimating the number of vehicles in traffic congestion \cite{DBLP:conf/eccv/MundhenkKSB16,DBLP:conf/eccv/Onoro-RubioL16,Zhang_2017_ICCV,Hsieh_2017_ICCV,Marsden_2018_CVPR}, counting the cells and bacteria from microscopic images~\cite{lempitsky2010learning,DBLP:journals/tmi/Sirinukunwattana16,DBLP:conf/eccv/BorstelKSRRH16,DBLP:conf/eccv/WalachW16,DBLP:conf/iccvw/CohenBGLB17}, and animal crowd estimations for ecological survey~\cite{Ma_2015_CVPR,DBLP:conf/eccv/ArtetaLZ16,Laradji_2018_ECCV}, to name a few.
	
	Crowd counting is a very challenging task because: 1) dense crowds often have heavy overlaps and occlusions between each other; 2) perspective effects may cause large variations in human size, shape, and appearance in the image. 
	In the past decade, a number of crowd counting algorithms \cite{DBLP:journals/tsmc/LinCC01,DBLP:conf/cvpr/ZhaoN03,DBLP:conf/icpr/LiZHT08,DBLP:conf/cvpr/GeC09,DBLP:conf/cvpr/ChanLV08,DBLP:conf/dicta/RyanDFS09,lempitsky2010learning,DBLP:conf/icpr/FiaschiKNH12,DBLP:conf/cvpr/ChenGXL13,DBLP:conf/iccv/PhamKYO15} have been proposed in the literature. Recently, crowd counting methods using Convolutional Neural Networks (CNNs) have made remarkable progresses~\cite{Zhang_2015_CVPR,DBLP:conf/mm/WangZYLC15,DBLP:conf/icip/ShangAB16,Chattopadhyay_2017_CVPR,Zhang_2016_CVPR,DBLP:conf/eccv/ZhaoLZW16,Shi_2018_CVPR,Deb_2018_CVPR_Workshops,Cao_2018_ECCV,Ranjan_2018_ECCV,Idrees_2018_ECCV}.
	The best performing methods are mostly based on the density map estimation, which typically obtain the crowd count by predicting a density map for the input image and then summing over the estimated density map. Nowadays, publicly available datasets~\cite{DBLP:conf/cvpr/IdreesSSS13,Zhang_2016_CVPR,Idrees_2018_ECCV} for training crowd count estimators only provide point annotations for each training image, \ie, only one pixel of each person is labeled (typically the center of the head). Currently, the most common approach for using these annotations is to first convert the point annotations for each training image to a ``ground-truth'' density map using the Gaussian kernel, and then train a CNN model by regressing the value at each pixel in this density map. With such pixel-level strict supervisions, the accuracy of a CNN model is highly dependent on the quality of the obtained ``ground-truth'' density maps.
	
	Obviously, ``ground-truth'' density maps obtained by applying a hypothetical Gaussian kernel to the point annotations can hardly be of top quality, 
	due to the occlusions, irregular crowd distributions, large variations in object size, shape, density, \etc.
	On the contrary, we propose Bayesian loss, which constructs a density contribution probability model from the point annotations. Then the expected count at each annotated point is calculated by summing the product of the contribution probability and estimated density at each pixel, which can be reliably supervised by the ground-truth count value (apparently, \emph{one}).
	Compared with previous loss functions that constrain the density value at every pixel, our proposed training loss supervises on the count expectation at each annotated point, instead.
	
	Extensive experimental evaluations show that the proposed loss function substantially outperforms the baseline training loss on UCF-QNRF~\cite{Idrees_2018_ECCV}, ShanghaiTech~\cite{Zhang_2016_CVPR}, and UCF\_CC\_50~\cite{DBLP:conf/cvpr/IdreesSSS13} benchmark datasets. Moreover, our proposed loss function equipped with the standard VGG-19 network~\cite{DBLP:journals/corr/SimonyanZ14a} as backbone, without using any external detectors or multi-scale architectures, achieves the state-of-the-art performances on all the benchmark datasets, especially with a magnificent improvement on the UCF-QNRF dataset compared to other methods.
	
	\section{Related Work}
	
	We review related works in the literature on crowd count estimation from the following respects.
	
	{\flushleft \textbf{Detection-then-counting.}} 
	Most of early works~\cite{DBLP:journals/tsmc/LinCC01,DBLP:conf/cvpr/ZhaoN03,DBLP:conf/icpr/LiZHT08,DBLP:conf/cvpr/GeC09} estimate crowd count by detecting or segmenting individual objects in the scene.
	This kind of methods has to tackle great challenges from two respects.
	Firstly, they produce more accurate results (\eg bounding-boxes or masks of instances) than the overall count which is computational expensive and mostly suitable in lower density crowds. In overcrowded scenes, clutters and severe occlusions make it unfeasible to detect every single person, despite the progresses in related fields~\cite{lazebnik2006beyond,felzenszwalb2008discriminatively,wang2010locality,NIPS2012_4824,ren2015faster,Stewart_2016_CVPR,Zhou_2017_CVPR,Wei_2018_CVPR,Wang_2018_CVPR,Ren_2018_CVPR,Wei_2018_ECCV,Zhang_2018_ECCV,Zhou_Wang_Meng:2019,Xu_2019_ICCV}.
	Secondly, training object detectors require bounding-box or instance mask annotations, which is much more labor-intensive 
	in dense crowds. Thus most of current counting datasets only provide a one-point label per object.
	
	{\flushleft \textbf{Direct count regression.}} 
	To avoid the more complex detection problem, some researchers proposed to directly learn a mapping from image features to their counts~\cite{DBLP:conf/cvpr/ChanLV08,DBLP:conf/dicta/RyanDFS09,DBLP:conf/cvpr/ChenGXL13,Liu_2015_ICCV,DBLP:conf/mm/WangZYLC15,DBLP:conf/icip/ShangAB16,Chattopadhyay_2017_CVPR}. 
	Former methods~\cite{DBLP:conf/cvpr/ChanLV08,DBLP:conf/dicta/RyanDFS09,DBLP:conf/cvpr/ChenGXL13} in this category rely on hand-crafted features, such as SIFT, LBP \etc, and then learn a regression model.
	Chan \etal~\cite{DBLP:conf/cvpr/ChanLV08} proposed to extract edge, texture and other low-level features of the crowds, and lean a Gaussian Process regression model for crowd counting.
	Chen \etal~\cite{DBLP:conf/cvpr/ChenGXL13} proposed to transform low-level image features into a cumulative attribute space where each dimension has clearly defined semantic interpretation that captures how the crowd count value changes continuously and cumulatively.
	Recent methods~\cite{DBLP:conf/mm/WangZYLC15,DBLP:conf/icip/ShangAB16,Chattopadhyay_2017_CVPR} resort to deep CNNs for end-to-end learning. Wang \etal~\cite{DBLP:conf/mm/WangZYLC15} adopted an AlexNet architecture where the final fully connected layer of 4096 neurons is replaced by a single neuron for predicting the scalar count value. Shang \etal~\cite{DBLP:conf/icip/ShangAB16} proposed to extract a set of high level image features via a CNN firstly, and then map the features to local counts using a Long Short-Term Memory (LSTM) unit.
	These direct regression methods are more efficient than detection based methods, however, they do not fully utilized available point supervisions.
	
	{\flushleft \textbf{Density map estimation.}} 
	This kind of methods~\cite{lempitsky2010learning,DBLP:conf/icpr/FiaschiKNH12,DBLP:conf/iccv/PhamKYO15} take advantage of the location information to learn a map of density values for each training sample and the final count estimation can be obtained by summing over the predicted density map.
	Lempitsky and Zisserman~\cite{lempitsky2010learning} proposed to transform the point annotations into a density map by the Gaussian kernel as ``ground-truth''. Then they train their models using a least-square objective.
	This kind of training framework has been widely used in recent methods~\cite{DBLP:conf/icpr/FiaschiKNH12,DBLP:conf/iccv/PhamKYO15}.
	Furthermore, thanks to the excellent feature learning ability of deep CNNs, CNN based density map estimation methods~\cite{Zhang_2015_CVPR,Zhang_2016_CVPR,Xiong_2017_ICCV,Shi_2018_CVPR,Liu_2018_CVPR,DBLP:conf/ijcai/LiuWLOL18,Cao_2018_ECCV,Ranjan_2018_ECCV,Idrees_2018_ECCV} have achieved the state-of-the-art performance for crowd counting.
	One major problem of this framework is how to determine the optimal size of the Gaussian kernel which is influenced by many factors.
	To make matters worse, the models are trained by a loss function which applies supervision in a pixel-to-pixel manner. Obviously, the performance of such methods highly depend on the quality of the generated ``ground-truth'' density maps.
	
	{\flushleft \textbf{Hybrid training.}} 
	Several works observed that crowd counting benefits from mixture training strategies, \eg, multi-task, multi-loss, \etc.
	Liu \etal~\cite{Liu2018} proposed DecideNet to adaptively decide whether to use a detection model or a density map estimation model. This approach takes the advantage of mixture-of-experts where a detection based model can estimate crowds accurately in low density scenes while the density map estimation model is good at handling crowded scenes.
	However, this method requires external pre-trained human detection models and is less efficient.
	Some researchers proposed to combine multiple losses to assist each other.
	Zhang \etal~\cite{Zhang_2015_CVPR} proposed to train a deep CNN by alternatively optimizing a pixel-wise loss function and a global count regression loss. A similar training approach was adopted by Zhang \etal~\cite{DBLP:conf/wacv/ZhangSC18}, in which they first train their model via the density map loss and then add a relative count loss in the last few epochs.
	Idrees \etal~\cite{Idrees_2018_ECCV} proposed a composition loss, which consists of $\ell_1$, $\ell_2$, and $\ell_{\infty}$ norm losses for the density map and a count regression loss. Compared to these hybrid losses, our proposed single loss function is simpler and more effective.
	
	\section{The Proposed Method}\label{sec:method}
	
	\subsection{Background and Motivation}\label{sec:motivation}
	
	Let $\{{\bf{D}}({\bf{x}}_m)>=0: m=1,2,\ldots,M\}$ be a density map, where ${\bf{x}}_m$ denotes a 2D pixel location, and $M$ is the number of pixels in the density map. 
	Let $\{({\bf{z}}_n,y_n): n=1,2,\ldots ,N\}$ denote the point annotation map for a sample image, where $N$ is the total crowd count, ${\bf{z}}_n$ is a head point position and $y_n=n$ is the corresponding label. The point annotation map contains only one pixel for each person (typically the center of the head), which is sparse, and contains no information about the object size and shape. It is difficult to directly use such point annotation maps to train a density map estimator.
	A common remedy to this difficulty is to convert it to a ``ground-truth'' density map using the Gaussian kernel.
	\begin{equation}
	\begin{aligned}
	{\bf{D}}^{gt}({\bf{x}}_m) \overset{\underset{\mathrm{def}}{}}{=}~& \sum\limits_{n=1}^N \mathcal{N}({{\bf{x}}_m};{{\bf{z}}_n},\sigma^2{{\bf{1}}_{2 \times 2}}) \\
	=~& \sum\limits_{n=1}^N \frac{1}{{\sqrt {2\pi } {\sigma}}}\exp ( - \frac{{\left\| {{{\bf{x}}_m} - {{\bf{z}}_n}} \right\|_2^2}}{{2\sigma^2}}),
	\end{aligned}
	\label{eq:gt_density_map}
	\end{equation}
	where ${\mathcal{N}({\bf{x}}_m;{{\bf{z}}_n},\sigma ^2{{\bf{1}}_{2 \times 2}})}$ denotes a 2D Gaussian distribution evaluated at ${\bf{x}}_m$, with the mean at the annotated point ${\bf{z}}_n$, and an isotropic covariance matrix $\sigma^2{{\bf{1}}_{2 \times 2}}$. 
	
	Many recent works use the above ``ground-truth'' density map as the learning target, and train a density map estimator using the following loss function:
	\begin{equation}
	\mathcal{L}^{baseline} = \sum\limits_{m = 1}^M{\mathcal{F}\left({\bf{D}}^{gt}({\bf{x}}_m) - {\bf{D}}^{est}({\bf{x}}_m)\right)},
	\label{eq:previous_loss}
	\end{equation}
	where $\mathcal{F}(\cdot)$ is a distance function and $D^{est}$ is the estimated density map. 
	If a fix-sized Gaussian kernel is adopted $\sigma \overset{\underset{\mathrm{def}}{}}{=} const$, it is assumed that all the people in a dataset have the same head size and shape, which is obviously not  true due to the occlusions, irregular crowd distributions, perspective effects, \etc. An alternative solution is to use an adaptive Gaussian kernel~\cite{Zhang_2016_CVPR,Idrees_2018_ECCV} for each $n$: $\sigma_n \propto d_n$, where $d_n$ is a distance that depends on its nearest neighbors in the spatial domain, which assumes that the crowd is evenly distributed.
	Some other methods utilize specific information such as camera parameters to get a more accurate perspective map, but in general such information is not available.
	
	We argue that the point annotations in the available crowd counting datasets can rather be considered as weak labels for density map estimation. It is more reasonable to take such annotations as priors or likelihoods instead of the learning targets. Loss functions that impose such strict, pixel-to-pixel supervisions as Eq.~\eqref{eq:previous_loss} on density map are not always beneficial to enhance the count estimation accuracy when used to train a CNN model, because it forces the model to learn inaccurate, or even erroneous information.
	
	\subsection{Bayesian Loss}\label{sec:loss}
	
	Let ${\bf{x}}$ be a random variable that denotes the spatial location and $y$ be a random variable that represents the annotated head point. 
	Based on the above discussions, instead of converting point annotations into the ``ground-truth'' density maps generated by Eq.~\eqref{eq:gt_density_map} as the learning targets, we propose to construct likelihood functions of ${\bf{x}}_m$ given label $y_n$ from them,
	\begin{equation}
	p({\bf{x}}={{\bf{x}}_m}|y=y_{n}) = \mathcal{N}({{\bf{x}}_m};{{\bf{z}}_n},\sigma^2{{\bf{1}}_{2 \times 2}}).
	\label{eq:foreground_likelihood}
	\end{equation}
	To simplify the notations, we omit the random variable ${\bf{x}}$ and $y$ in the following formulations, \eg, Eq.~\eqref{eq:foreground_likelihood} becomes $p({{\bf{x}}_m}|y_{n}) = \mathcal{N}({{\bf{x}}_m};{{\bf{z}}_n},\sigma^2{{\bf{1}}_{2 \times 2}})$.
	According to Bayes' theorem, given a pixel location ${\bf{x}}_m$ in the density map, the posterior probability of ${\bf{x}}_m$ having the label $y_n$ can be computed as:
	\begin{equation}
	\begin{aligned}
	p({y_n}|{{\bf{x}}_m}) &= \frac{{p({{\bf{x}}_m}|{y_n})p({y_n})}}{{p({{\bf{x}}_m})}} = \frac{{p({{\bf{x}}_m}|{y_n})p({y_n})}}{{\sum\limits_{n = 1}^N {p({{\bf{x}}_m}|{y_n})p({y_n})} }} \\
	&= \frac{{p({{\bf{x}}_m}|{y_n})}}{{\sum\limits_{n = 1}^N {p({{\bf{x}}_m}|{y_n})} }} = \frac{\mathcal{N}({{\bf{x}}_m};{{\bf{z}}_n},\sigma^2{{\bf{1}}_{2 \times 2}})}{\sum\limits_{n = 1}^N \mathcal{N}({{\bf{x}}_m};{{\bf{z}}_n},\sigma^2{{\bf{1}}_{2 \times 2}})}\;.
	\end{aligned}
	\end{equation}
	In the above derivation, the third equality holds as we assume the equal prior probability $p(y_n)$ for each class label $y_n$, \ie $p(y_n) = \frac{1}{N}$, without loss of generality.
	In practice, if we know the prior that crowds are more or less tend to appear at certain places, a tailored $p(y_n)$ can be applied here.
	
	With the posterior label probability  ${p(y_n|{\bf{x}}_m)}$ and estimated density map ${\bf{D}}^{est}$, we derive the Bayesian loss as follows. let $c_n^m$ denotes the count that ${\bf{x}}_m$ contributes to $y_n$, and $c_n$ be the total count associated with $y_n$, we have the expectation of $c_n$ as:
	\begin{equation}
	\begin{aligned}
	E[c_n] &= E[\sum\limits_{m = 1}^M c_n^m] = \sum\limits_{m = 1}^M E[c_n^m] \\
	&= \sum\limits_{m = 1}^M {p({y_n}|{{\bf{x}}_m}){{\bf{D}}^{est}}({{\bf{x}}_m})}.
	\end{aligned}
	\end{equation}
	Obviously, the ground-truth count $c_n$ at each annotation point is one, therefore we have the following loss function:
	\begin{equation}
	\mathcal{L}^{Bayes} = \sum\limits_{n = 1}^N \mathcal{F}({1 - E[c_n]}),
	\end{equation}
	where $\mathcal{F}(\cdot)$ is a distance function and we adopt $\ell_1$ distance in our experiments.
	A special case should be handled when there is no object in a training image. In such scenario we directly force the sum of the density map to zero. Our proposed loss function is differentiable and can be readily applied to a given CNN using the standard back propagation training algorithm.
	
	At the inference stage, we do not have to know the posterior label probability $p(y_n|\bf{x}_m)$ in advance, because when we sum over the estimated density map, we eliminate $p(y_n|\bf{x}_m)$ as follows:
	\begin{equation}
	\begin{aligned}
	C = \sum\limits_{n = 1}^N {E[c_n]} &= \sum\limits_{n = 1}^N {\sum\limits_{m = 1}^M {{{p(y_n|{\bf{x}}_m)}}{{\bf{D}}^{est}}({{\bf{x}}_m})} } \\
	&= \sum\limits_{m = 1}^M {\sum\limits_{n = 1}^N {{{p(y_n|{\bf{x}}_m)}}{{\bf{D}}^{est}}({{\bf{x}}_m})} } \\
	&= \sum\limits_{m = 1}^M {{{\bf{D}}^{est}}({{\bf{x}}_m})}.
	\end{aligned}
	\end{equation}
	
	\subsection{Background Pixel Modelling}
	
\begin{figure}[!t]
		\centering
		\fbox{\includegraphics[width=0.98\linewidth]{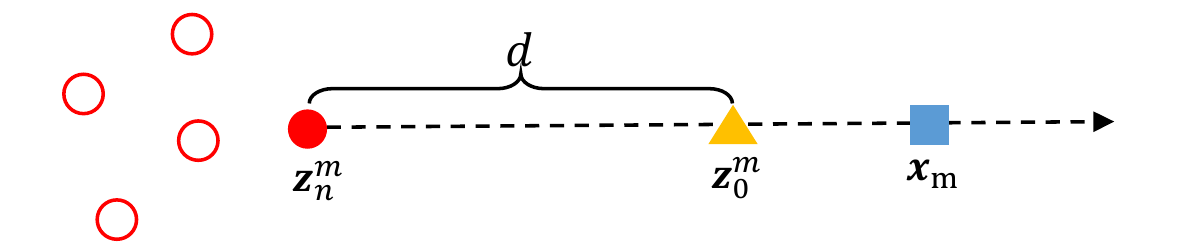}}
		\vspace{2mm}
		\caption{\emph{Geometrical illustration of the dummy background point, where ${\bf{x}}_m$ denotes a pixel in the density map, ${\bf{z}}_n^m$ is its nearest head point and ${\bf{z}}_0^m$ is the defined dummy background point.}}
		\label{fig:dummy_point}
	\end{figure}

	For background pixels that are far away from any of the annotation points, it makes no sense to assign them to any head label $y_n$. To better model the background pixels, we introduce an additional background label $y_0=0$, in addition to the head labels $\{y_n=n: n=1,2,\ldots ,N\}$. Then, the posterior label probability can be rewritten as:
	\begin{equation}
	\begin{aligned}
	p({y_n}|{{\bf{x}}_m}) &= \frac{{p({{\bf{x}}_m}|{y_n})p({y_n})}}{{\sum\limits_{n = 1}^N {p({{\bf{x}}_m}|{y_n})p({y_n}) + p({{\bf{x}}_m}|{y_0})p({y_0})} }} \\
	&= \frac{{p({{\bf{x}}_m}|{y_n})}}{{\sum\limits_{n = 1}^N {p({{\bf{x}}_m}|{y_n}) + p({{\bf{x}}_m}|{y_0})} }}.
	\end{aligned}
	\end{equation}
	The last equation is simplified with the assumption $p(y_n)=p(y_0) = \frac{1}{N+1}$, without loss of generality.
	Similarly, we have:
	\begin{equation}
	{p({y_0}|{{\bf{x}}_m}) = \frac{{p({{\bf{x}}_m}|{y_0})}}{{\sum\limits_{n = 1}^N {p({{\bf{x}}_m}|{y_n}) + p({{\bf{x}}_m}|{y_0})} }}}.
	\label{eq:background_posterior}
	\end{equation}
	And the expected counts for each person and for the entire background are defined as:
	\begin{equation}
	E[c_n] = \sum\limits_{m = 1}^M {p({y_n}|{{\bf{x}}_m}){{\bf{D}}^{est}}({{\bf{x}}_m})},
	\label{eq:foreground}
	\end{equation}
	
	\begin{equation}
	E[c_0] = \sum\limits_{m = 1}^M {p({y_0}|{{\bf{x}}_m}){{\bf{D}}^{est}}({{\bf{x}}_m})}.
	\label{eq:background}
	\end{equation}
	In this case, the summation over the whole density map $\sum\nolimits_{m = 1}^M{{{\bf{D}}^{est}}({{\bf{x}}_m})}$ consists of the foreground counts $\sum\nolimits_{n = 1}^N {E[c_n]}$ and the background count $E[c_0]$.
	Obviously, we would like the background count to be zero and the foreground count at each annotation point equals to one, thus we have the following enhanced loss function,
	
	\begin{equation}
	\begin{aligned}
	{\mathcal{L}^{Bayes+}} = \sum\limits_{n = 1}^N{\mathcal{F}({1-E[c_n]})} + \mathcal{F}(0-E[c_0]).
	\end{aligned}
	\end{equation}
	To define the background likelihood, we construct a dummy background point for each pixel,
	\begin{equation}
	{{\bf{z}}_{0}^{m}} = {{\bf{z}}_{n}^{m}}+d\frac{{\bf{x}}_m-{\bf{z}}_n^m}{{{\left\| {\left. {{{\bf{x}}_m} - {\bf{z}}_{n}^{m}} \right\|_2} \right.}}},
	\label{eq:dummy_point}
	\end{equation}
	where ${\bf{z}}_{n}^{m}$ denotes the nearest head point of ${\bf{x}}_{m}$, and $d$ is a parameter that controls the margin between the head and the dummy background points. As illustrated in Fig.~\ref{fig:dummy_point}, with the defined dummy background point ${\bf{z}}_{0}^{m}$, for a pixel ${\bf{x}}_{m}$ that is far away from head points, it can be assigned to the background label instead.
	Here we also use the Gaussian kernel to define the background likelihood,
	\begin{equation}
	\begin{aligned}
	p({{\bf{x}}_m}|{y_0}) \overset{\underset{\mathrm{def}}{}}{=}~& \mathcal{N}({{\bf{x}}_m};{{\bf{z}}_{0}^{m}},\sigma^2{{\bf{1}}_{2 \times 2}}) \\
	=~& \frac{1}{{\sqrt {2\pi } {\sigma}}}\exp ( - \frac{{(d - \left\| {{{\bf{x}}_m} - {{\bf{z}}_n^m}} \right\|_2)^2}}{{2\sigma^2}}).
	\end{aligned}
	\label{eq:background_likelihood}
	\end{equation}

	\setlength{\fboxsep}{0in}
	
	\begin{figure*}[!t]
		\centering
		\setlength{\tabcolsep}{2pt}
		\begin{tabular}{ccc}
			\includegraphics[width=0.315\linewidth]{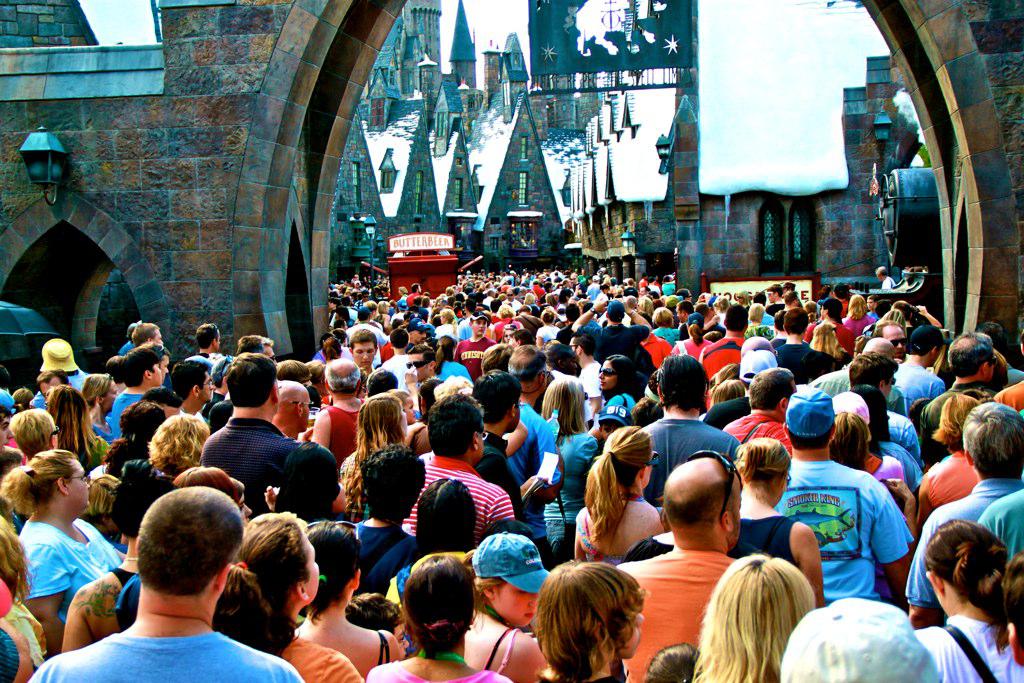}   & 
			\includegraphics[width=0.315\linewidth]{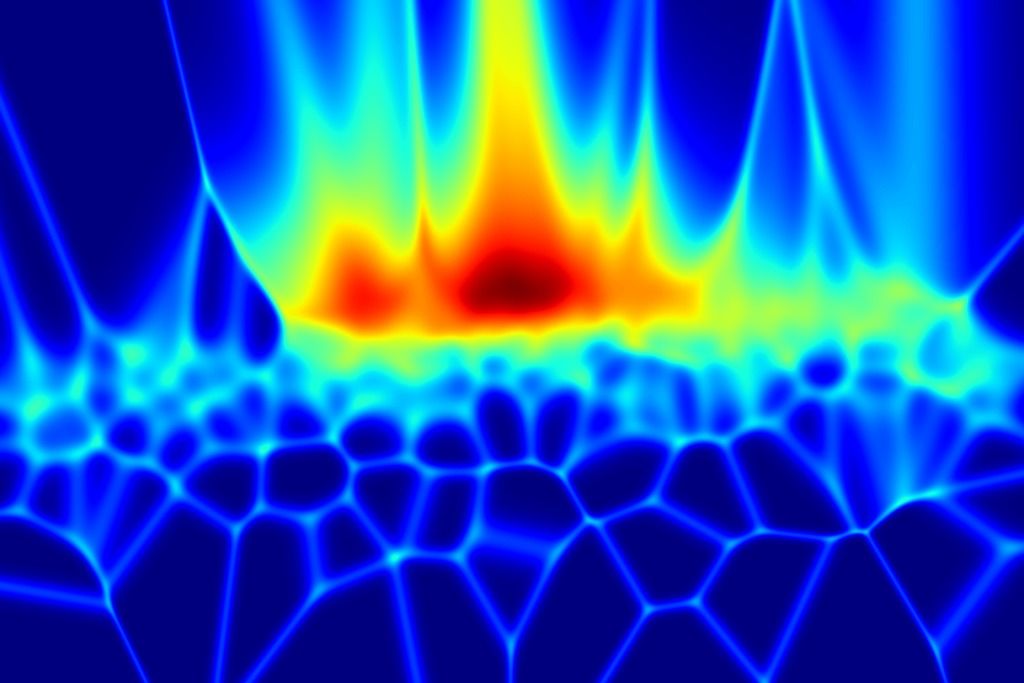} &
			\includegraphics[width=0.315\linewidth]{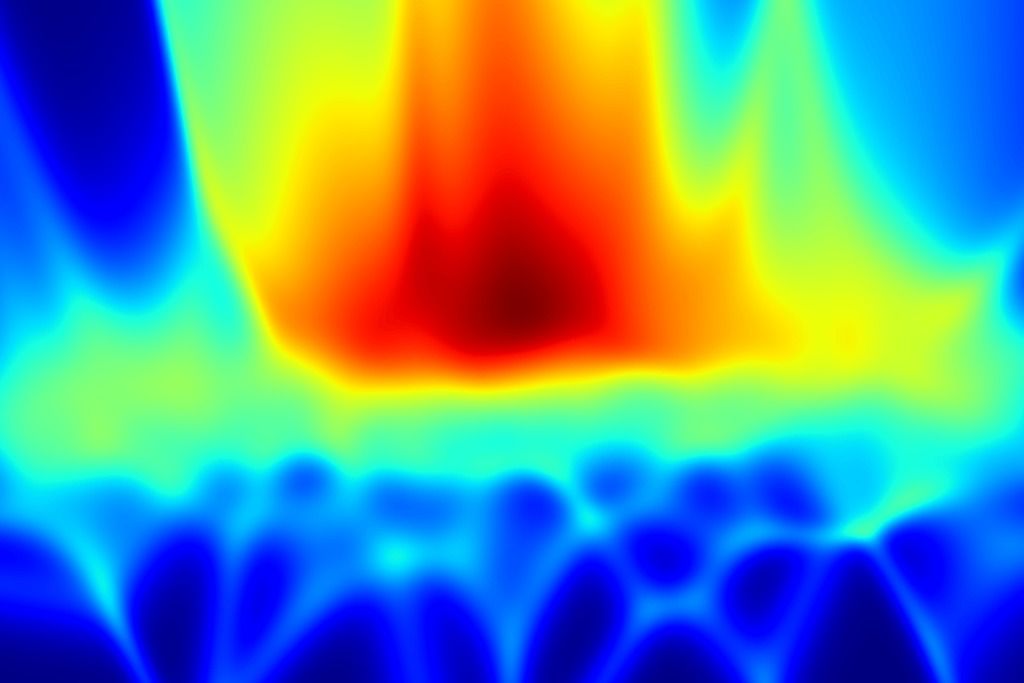} \\
			\small{(a) Input image} & \small{(b) Without $y_0$, $\sigma=16$} & \small{(c) Without $y_0$, $\sigma=32$} \\
			\includegraphics[width=0.315\linewidth]{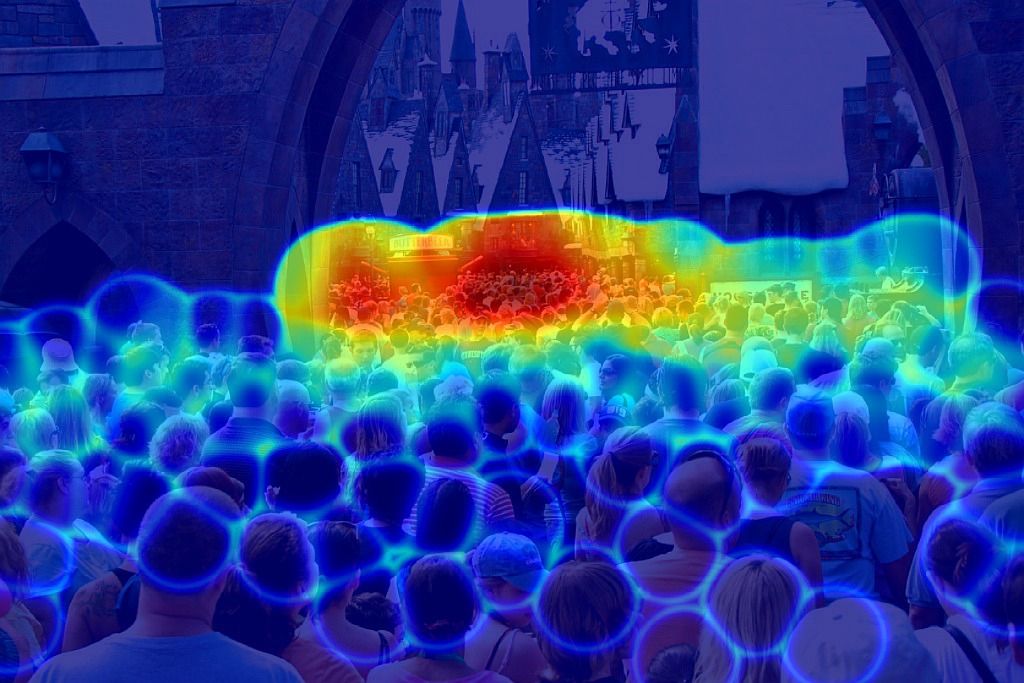} &
			\includegraphics[width=0.315\linewidth]{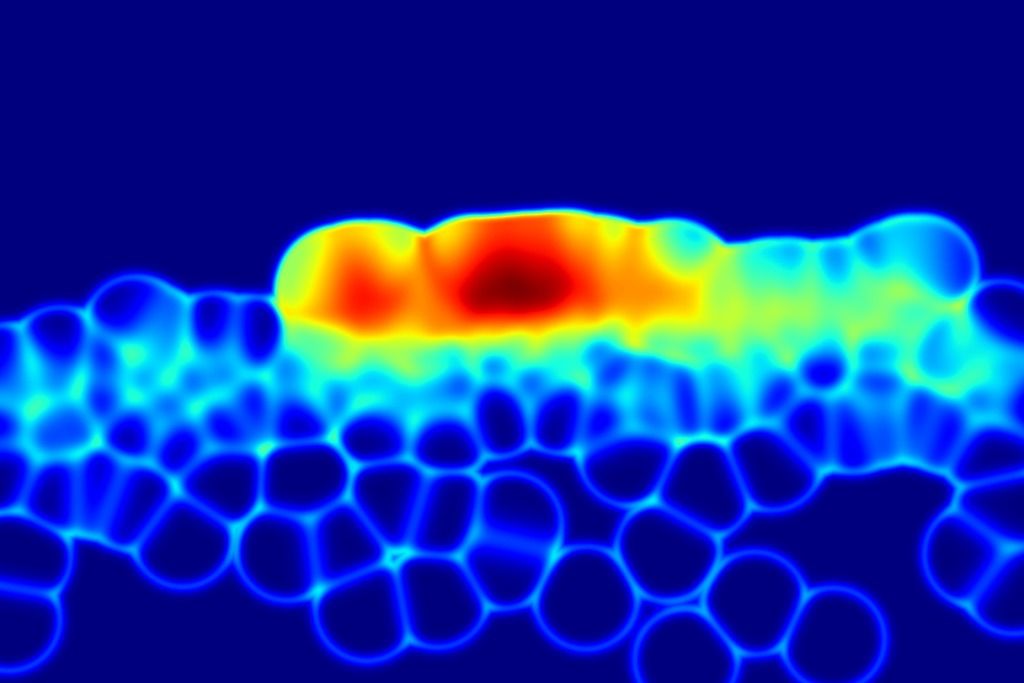} &
			\includegraphics[width=0.315\linewidth]{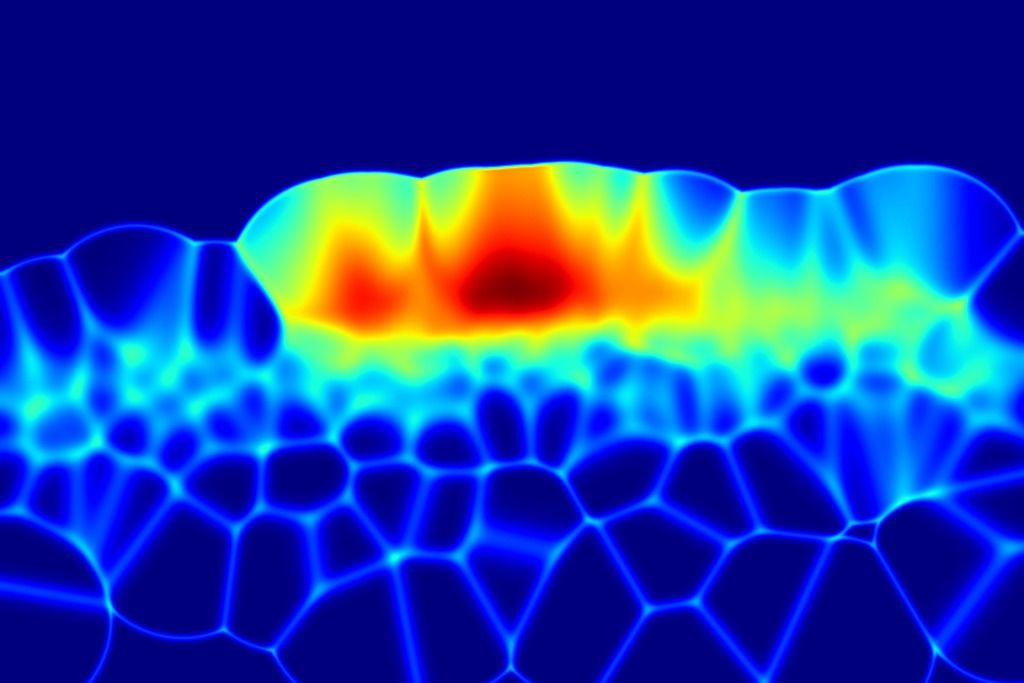} \\
			\small{(d) Blend of (a,e) } & \small{(e) With $y_0$, $\sigma=16,d=100$} & \small{(f) With $y_0$, $\sigma=16,d=200$}
		\end{tabular}
		\vspace{1mm}
		\caption{\emph{Visualization of the posterior label probability. We construct an entropy map using Eq.~\eqref{eq:entropy}, which measures the uncertainty on the label a pixel in the density map belongs to. The color is warmer, the value is larger. (a) Input image. (b)-(c): Entropy maps with different $\sigma$, without background pixel modelling. (e)-(f): Entropy maps with different $d$, with background pixel modelling. (d): Blend of the input image and the entropy map in (e).}}
		\label{fig:entropy}
	\end{figure*}
	
	\subsection{Visualization and Analysis}
	We build the entropy map ${\bf{E}}nt$ of label assignment for visualization and analysis, which is calculated for each pixel ${{\bf{x}}_m}$ as follows,
	\begin{equation}
	{\bf{E}}nt({{\bf{x}}_m}) =  - \sum\limits_{n = 0}^N {{p(y_n|{\bf{x}}_m)}\ln {p(y_n|{\bf{x}}_m)}}.
	\label{eq:entropy}
	\end{equation}
	The entropy measures the uncertainty on the label a pixel $\bf{x}_m$ in the density map belongs to.
	We display entropy maps with different settings in Fig.~\ref{fig:entropy} and have the following summarizes: 
	\begin{itemize}
		\setlength\itemsep{0em}
		\item The posterior could find the boundary between persons roughly.
		\item Dense crowd areas have higher entropy values than sparse areas.
		\item The parameter $\sigma$ controls the softness of the posterior label probability, comparing (b) and (c).
		\item Pixels far from crowds are handled better via background pixel modelling, comparing (b) and (e).
		\item The  parameter $d$ controls the margin between foreground and background, comparing (e) and (f).
	\end{itemize}
	
	\section{Experiments}
	
	\subsection{Evaluation Metrics}
	Crowd count estimation methods are evaluated by two widely used metrics: Mean Absolute Error (MAE) and Mean Squared Error (MSE), which are defined as follows:
	\begin{equation}
	MAE = \frac{{\rm{1}}}{K}\sum\limits_{k = 1}^K {\left| {N_k - C_k} \right|},
	\end{equation}
	\begin{equation}
	MSE = \sqrt {\frac{{\rm{1}}}{K}\sum\limits_{k = 1}^K {{{\left| {N_k - C_k} \right|}^2}} },
	\end{equation}
	where $K$ is the number of test images, $N_k$ and $C_k$ are the ground-truth count and the estimated count for the $k$-th image, respectively.
	
	\def\arraystretch{0.9}
	\setlength{\tabcolsep}{15.7pt}
	\begin{table*}[!t]
		\small
		\begin{center}
			\begin{tabular}{@{}l|cc|cc|cc|cc@{}}
				\toprule[1.0pt]
				Datasets &\multicolumn{2}{c|}{UCF-QNRF}&\multicolumn{2}{c|}{ShanghaiTechA}&\multicolumn{2}{c|}{ShanghaiTechB} &\multicolumn{2}{c}{UCF\_CC\_50}\\
				Methods & MAE & MSE & MAE & MSE & MAE & MSE & MAE & MSE\\
				\midrule[0.5pt]
				\textsc{Crowd-CNN}~\cite{Zhang_2015_CVPR} & - & - & 181.8 & 277.7 & 32.0 & 49.8 & 467.0 & 498.5\\
				\textsc{MCNN}~\cite{Zhang_2016_CVPR} & 277 & 426 & 110.2 & 173.2 & 26.4 & 41.3 & 377.6 & 509.1\\
				\textsc{CMTL}~\cite{DBLP:conf/avss/SindagiP17} & 252 & 514 & 101.3 & 152.4 & 20.0 & 31.1 & 322.8 & 341.4 \\
				\textsc{Switch-CNN}~\cite{Sam_2017_CVPR} & 228 & 445 & 90.4 & 135.0 & 21.6 & 33.4 & 318.1 & 439.2\\
				\textsc{CP-CNN}~\cite{Sindagi_2017_ICCV} & - & - & 73.6 & 106.4 & 20.1 & 30.1 & 295.8 & 320.9\\
				\textsc{ACSCP}~\cite{Shen_2018_CVPR} & - & - & 75.7 & 102.7 & 17.2 & 27.4 & 291.0 & 404.6\\
				\textsc{D-ConvNet}~\cite{Shi_2018_CVPR} & - & - & 73.5 & 112.3 & 18.7 & 26.0 & 288.4 & 404.7\\
				\textsc{IG-CNN}~\cite{Sam_2018_CVPR} & - & - & 72.5 & 118.2 & 13.6 & 21.1 & 291.4 & 349.4\\
				\textsc{ic-CNN}~\cite{Ranjan_2018_ECCV} & - & - & 68.5 & 116.2 & 10.7 & 16.0 & 260.9 & 365.5\\
				\textsc{SANet}~\cite{Cao_2018_ECCV} & - & - & 67.0 & 104.5 & 8.4 & 13.6 & 258.4 & 334.9 \\
				\textsc{CL-CNN}~\cite{Idrees_2018_ECCV} & 132 & 191 & - & - & - & - & - & - \\
				\midrule[0.5pt]
				\textsc{Baseline} & 106.8 & 183.7 & 68.6 & 110.1 & 8.5 & 13.9 & 251.6 & 331.3 \\
				Our \textsc{Bayesian} & 92.9  & 163.0 & 64.5 & 104.0 & 7.9 & 13.3 & 237.7 & 320.8 \\
				Our \textsc{Bayesian+} &\textbf{88.7}&\textbf{154.8}&\textbf{62.8}&\textbf{101.8}&\textbf{7.7}&\textbf{12.7}&\textbf{229.3}&\textbf{308.2}\\
				\bottomrule[1.0pt]		
			\end{tabular}
			\vspace{2mm}
			\caption{\emph{Benchmark evaluations on four benchmark crowd counting datasets using the MAE and MSE metrics. The baseline and our methods are trained using VGG-19.}}\label{tab:results}
		\end{center}
	\end{table*}
	
	\begin{figure*}[!t]
		\begin{center}
			\def\arraystretch{0.8}
			\setlength{\tabcolsep}{1pt}
			\begin{tabular}{cccccccc}
				\multicolumn{2}{c}{\small{GT Count: 909}} & \multicolumn{2}{c}{\small{Estimate: 1020.1}} & \multicolumn{2}{c}{\small{Estimate: 982.1}} & \multicolumn{2}{c}{\small{Estimate: 978.5}} \\
				\multicolumn{2}{c}{\includegraphics[width=0.239\linewidth]{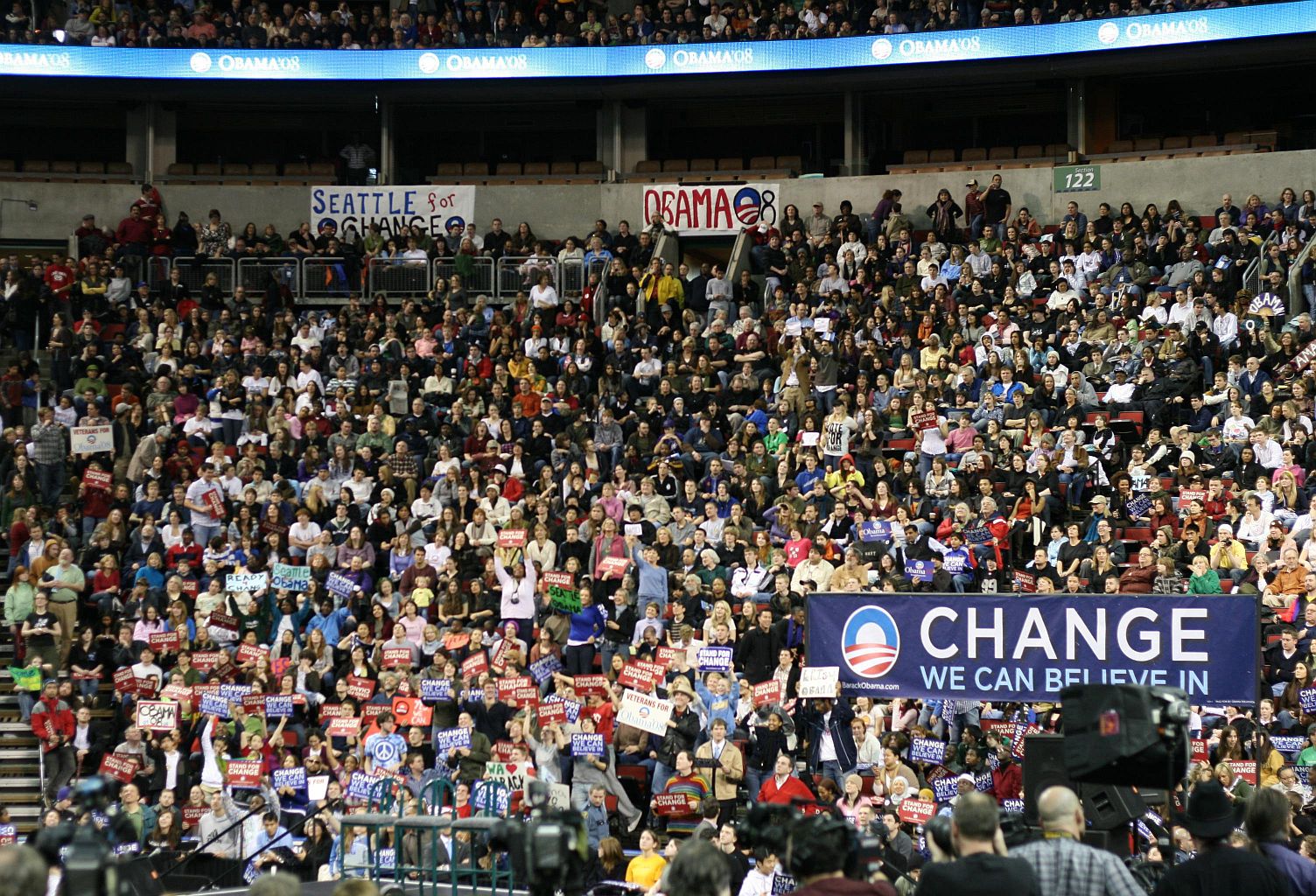}}  &
				\multicolumn{2}{c}{\includegraphics[width=0.239\linewidth]{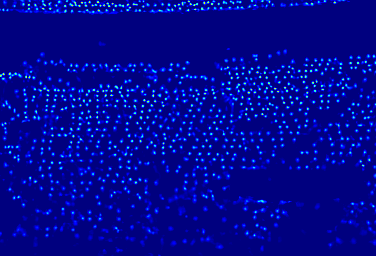}} &
				\multicolumn{2}{c}{\includegraphics[width=0.239\linewidth]{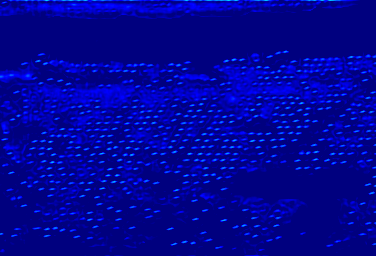}} &
				\multicolumn{2}{c}{\includegraphics[width=0.239\linewidth]{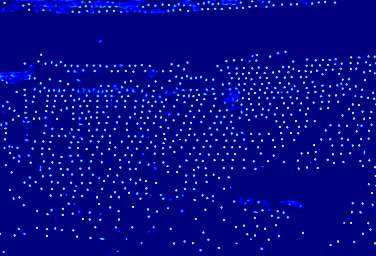}} \\
				\multicolumn{2}{c}{\small{GT Count: 2745}} & \multicolumn{2}{c}{\small{Estimate: 2452.3}} & \multicolumn{2}{c}{\small{Estimate: 2455.5}} & \multicolumn{2}{c}{\small{Estimate: 2470.3}} \\
				\multicolumn{2}{c}{\includegraphics[width=0.239\linewidth]{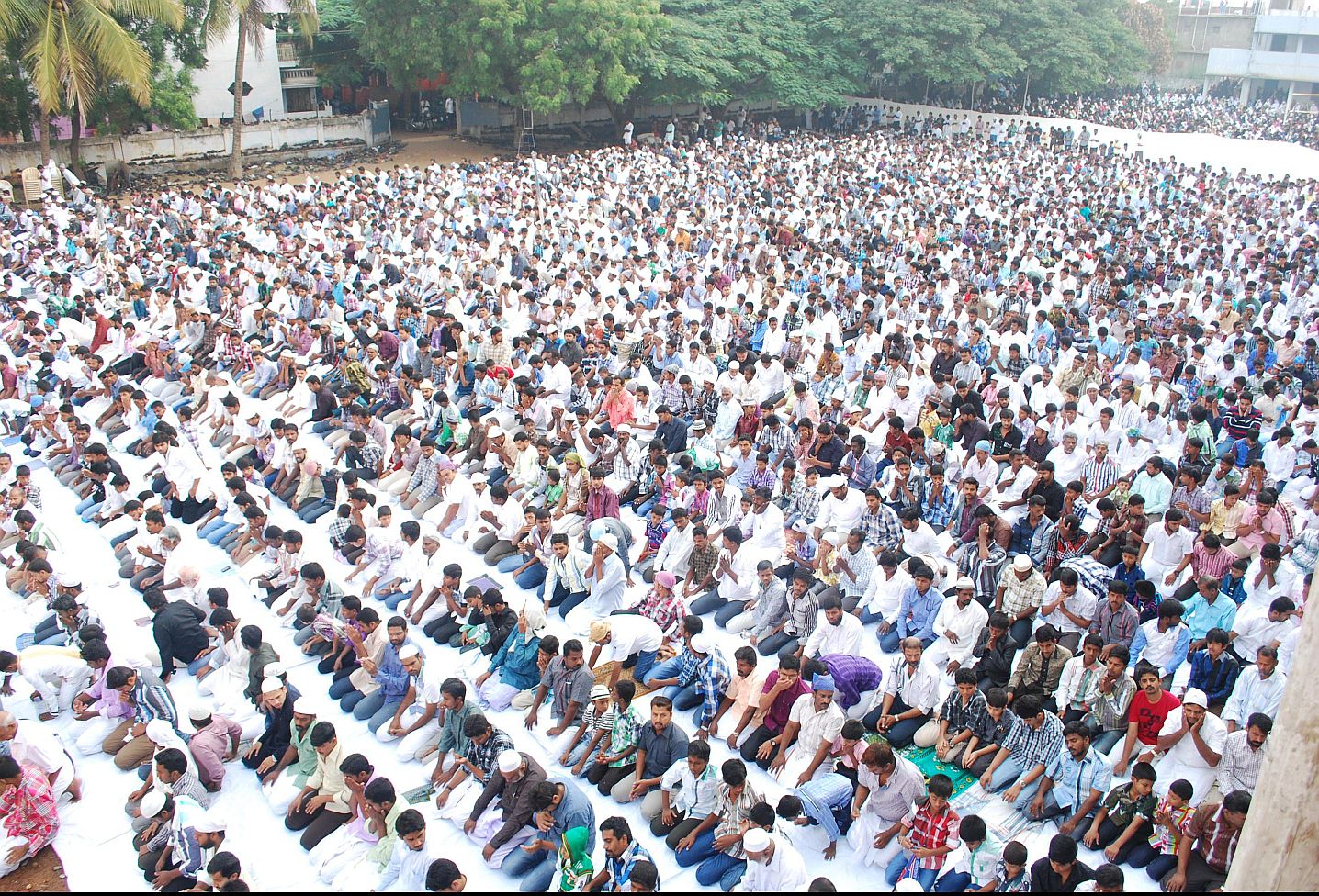}}  &
				\multicolumn{2}{c}{\includegraphics[width=0.239\linewidth]{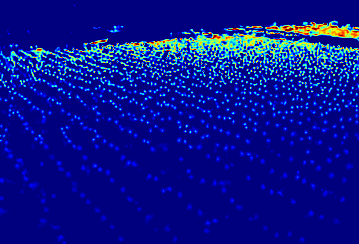}} &
				\multicolumn{2}{c}{\includegraphics[width=0.239\linewidth]{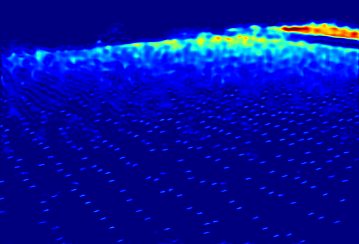}} &
				\multicolumn{2}{c}{\includegraphics[width=0.239\linewidth]{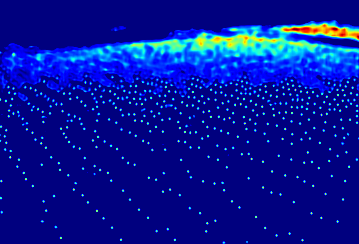}} \\
				\includegraphics[width=0.117\linewidth]{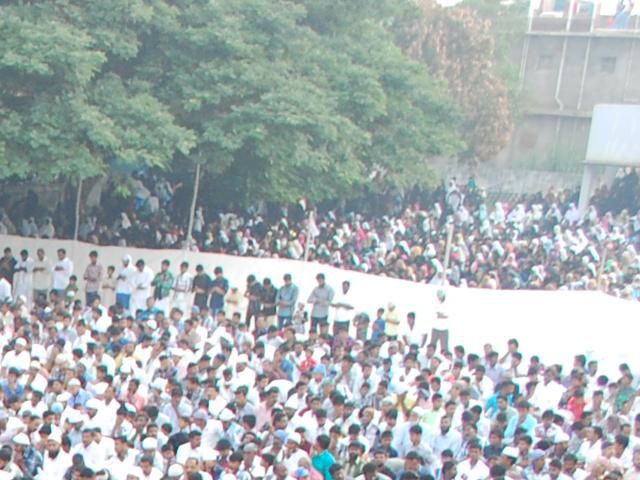}   &
				\includegraphics[width=0.117\linewidth]{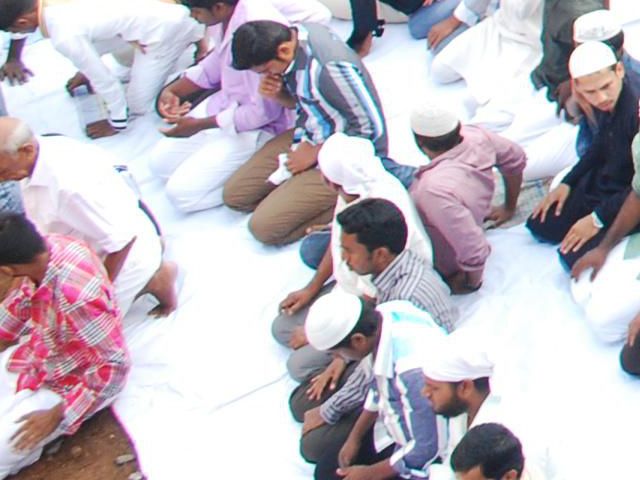}   &
				\includegraphics[width=0.117\linewidth]{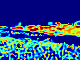} &
				\includegraphics[width=0.117\linewidth]{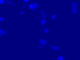} &
				\includegraphics[width=0.117\linewidth]{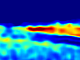} &
				\includegraphics[width=0.117\linewidth]{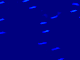} &
				\includegraphics[width=0.117\linewidth]{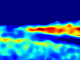} &
				\includegraphics[width=0.117\linewidth]{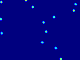} \\
				\multicolumn{2}{c}{\small{GT Count: 1616}} & \multicolumn{2}{c}{\small{Estimate: 1946.7}} & \multicolumn{2}{c}{\small{Estimate: 1686.7}} & \multicolumn{2}{c}{\small{Estimate: 1602.3}} \\
				\multicolumn{2}{c}{\includegraphics[width=0.239\linewidth]{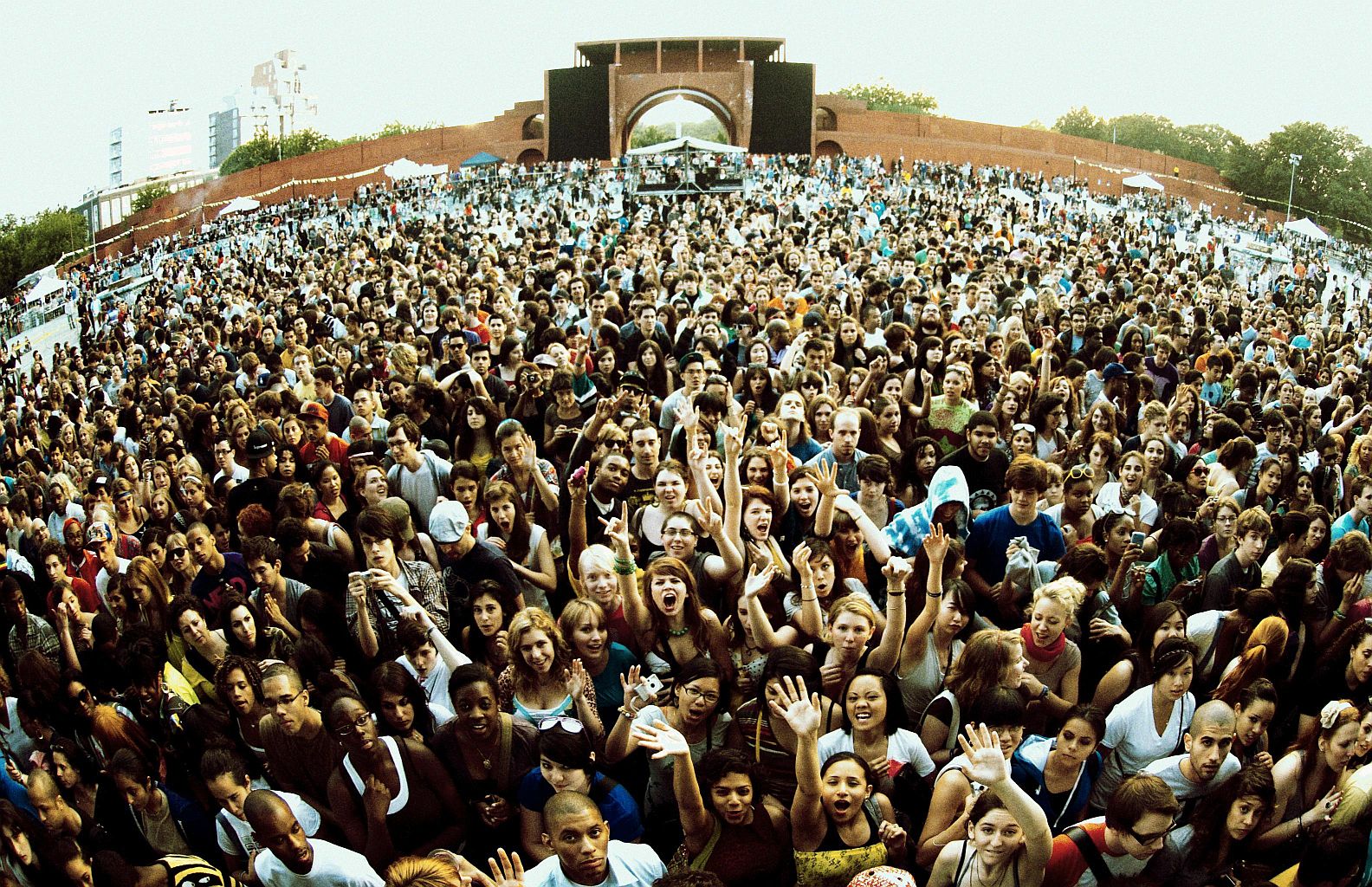}}  &
				\multicolumn{2}{c}{\includegraphics[width=0.239\linewidth]{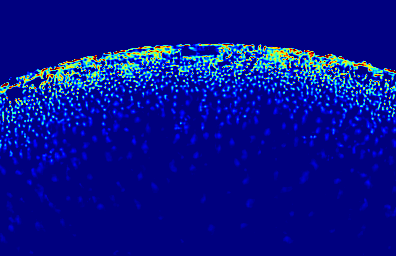}} &
				\multicolumn{2}{c}{\includegraphics[width=0.239\linewidth]{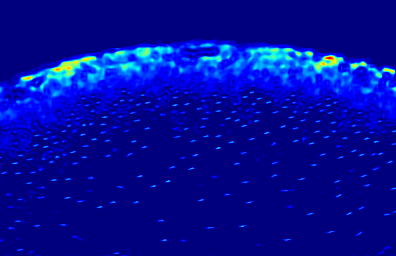}} &
				\multicolumn{2}{c}{\includegraphics[width=0.239\linewidth]{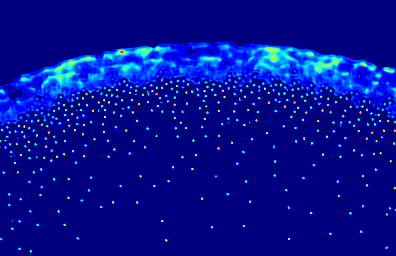}} \\
				\includegraphics[width=0.117\linewidth]{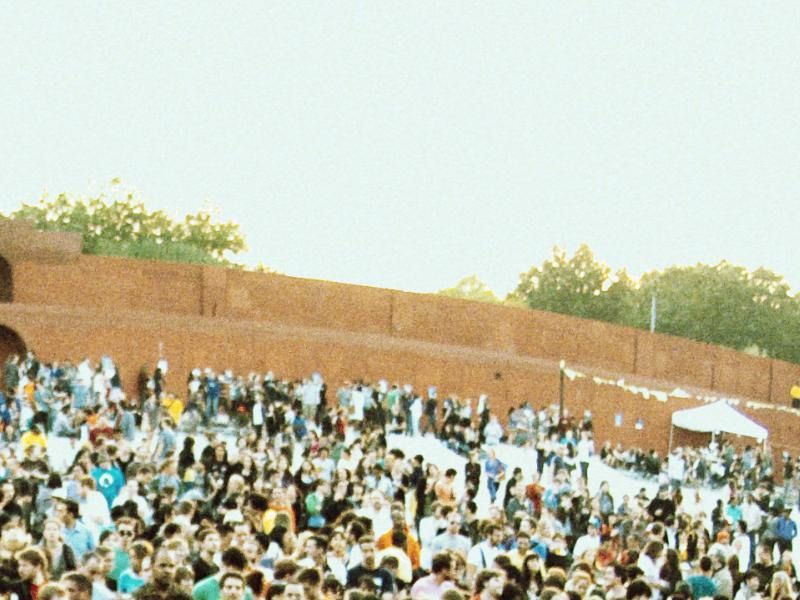}   &
				\includegraphics[width=0.117\linewidth]{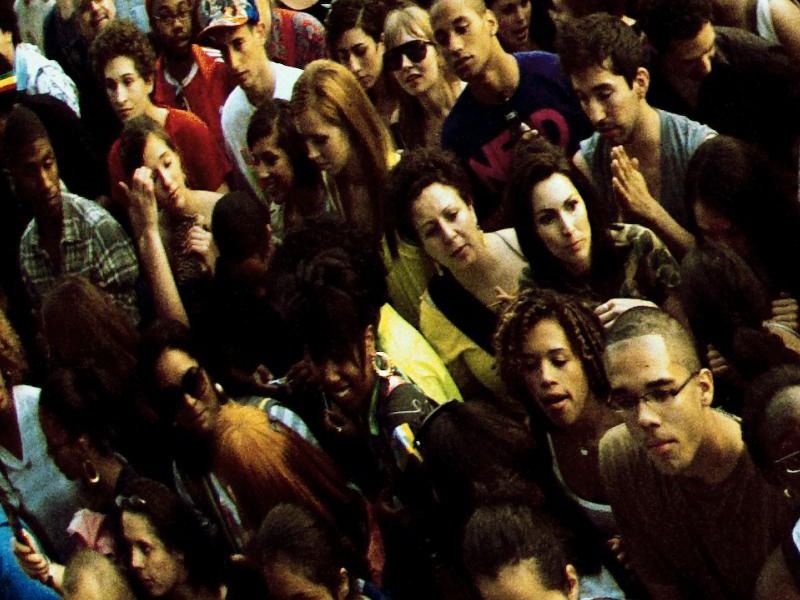}   &
				\includegraphics[width=0.117\linewidth]{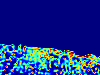} &
				\includegraphics[width=0.117\linewidth]{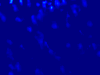} &
				\includegraphics[width=0.117\linewidth]{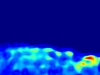} &
				\includegraphics[width=0.117\linewidth]{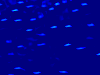} &
				\includegraphics[width=0.117\linewidth]{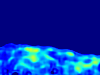} &
				\includegraphics[width=0.117\linewidth]{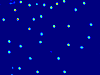} \\
				\multicolumn{2}{c}{\small{(a) Input image}} & \multicolumn{2}{c}{\small{(b) \textsc{Baseline}}} & \multicolumn{2}{c}{\small{(c) Our \textsc{Bayesian}}} & \multicolumn{2}{c}{\small{(d) Our \textsc{Bayesian+}}} \\
			\end{tabular}
			\vspace{1mm}
			\caption{\small{\emph{Density maps generated by (b) \textsc{Baseline}, our (c) \textsc{Bayesian}, and (d) \textsc{Bayesian+}. The color is warmer, the density is higher. Note that in dense crowds, \textsc{Baseline} often produces abnormal values, while in sparse areas, it can not localize each person well. In contrast, our methods give more accurate count estimate and localization, respectively.}}}
			\label{fig:viz}	
		\end{center}
	\end{figure*}
	
	\subsection{Datasets}\label{sec:dataset}
	Experimental evaluations are conducted using four widely used crowd counting benchmark datasets: UCF-QNRF~\cite{Idrees_2018_ECCV}, UCF\_CC\_50~\cite{DBLP:conf/cvpr/IdreesSSS13}, ShanghaiTech~\cite{Zhang_2016_CVPR} part A and part B. These datasets are described as follows. 
	
	{\flushleft \textbf{UCF-QNRF}~\cite{Idrees_2018_ECCV}} 
	is the latest and largest crowd counting dataset including 1535 images crawled from Flickr with 1.25 million point annotations. It is a challenging dataset because it has a wide range of counts, image resolutions, light conditions and viewpoints. The training set has 1,201 images and the remaining 334 images are used for testing.
	
	{\flushleft \textbf{ShanghaiTech}~\cite{Zhang_2016_CVPR}}
	consists of part A and part B. In part A, there are 300 images for training and 182 images for testing. All the images are crawled from the Internet, and most of them are images of very crowded scenes such as rallies and large sport events. Part B has 400 training images and 316 testing images captured from busy streets in Shanghai. Part A has a significantly higher density than part B.
	
	{\flushleft \textbf{UCF\_CC\_50}~\cite{DBLP:conf/cvpr/IdreesSSS13}} 
	contains 50 gray images with different resolutions. The average count for each image is 1,280, and the minimum and maximum counts are 94 and 4,532, respectively. 
	Since this is a small-scale dataset and no data split is defined for training and testing, we perform five-fold cross validations to get the average test result.
	
	\subsection{Implementation Details}
	{\flushleft \textbf{Network structure.}}
	We use a standard image classification network as our backbone, with the last pooling and the subsequent fully connected layers removed. In our experiments, we test two networks which are VGG-19~\cite{DBLP:journals/corr/SimonyanZ14a} and AlexNet~\cite{NIPS2012_4824}. 
	We upsample the output of the backbone to $1/8$ of the input image size by bilinear interpolation, and then feed it to a regression header, which consists of two $3 \times 3$ convolutional layers with 256 and 128 channels respectively, and a $1 \times 1$ convolutional layer, to get the density map. The regression header is initialized by the MSRA initializer~\cite{DBLP:conf/iccv/HeZRS15} and the backbone is pre-trained on ImageNet. The Adam optimizer with an initial learning rate $10^{-5}$ is used to update the parameters.
	
	{\flushleft \textbf{Training details.}}
	We augment the training data using random crop and horizontal flipping. We note that image resolutions in UCF-QNRF vary widely from 0.08 to 66 megapixels. However, a regular CNN can not deal with images with all kinds of scales due to its limited receptive field. Therefore, we limit the shorter side of each image within 2048 pixels in UCF-QNRF. Images are then randomly cropped for training, the crop size is $256 \times 256$ for ShanghaiTechA and UCF\_CC\_50 where image resolutions are smaller, and $512 \times 512$ for ShanghaiTechB and UCF-QNRF. We set the Gaussian parameter $\sigma$ in Eqs.~\eqref{eq:foreground_likelihood} and \eqref{eq:background_likelihood} to 8 and the distance parameter $d$ in Eq.~\eqref{eq:dummy_point} to 15\% of the shorter side of image. The parameters are selected on a validation set (120 images randomly sampled from the training set) of UCF-QNRF.
	
	\subsection{Experimental Evaluations}
	{\flushleft \textbf{Quantitative results.}}
	We compare our proposed method with the baseline and the state-of-the-art methods on the benchmark datasets described in Sec.~\ref{sec:dataset}.
	To make a fair comparison, the baseline method (\textsc{Baseline}) shares the same network structure (VGG-19) and training process as ours.
	We use Eq.~\eqref{eq:gt_density_map} to generate the ``ground-truth" density maps for the baseline method and follow previous works~\cite{Zhang_2016_CVPR,Idrees_2018_ECCV} to select parameters for the Gaussian kernel. Specifically, the geometry-adaptive kernels are adopted for UCF-QNRF, ShanghaiTechA and UCF\_CC\_50, while a fixed Gaussian kernel with $\sigma=15$ is used for ShanghaiTechB.
	We study both our basic Bayesian loss (\textsc{Bayesian}) and the enhanced Bayesian loss with the background pixel modelling (\textsc{Bayesian+}).
	We show the experimental results in Table~\ref{tab:results} and the highlights can be summarized as follows:
	\begin{itemize}
		\setlength\itemsep{0em}
		\item \textsc{Bayesian+}  achieves  the state-of-the-art  accuracy on all the four benchmark datasets. On the latest and the toughest UCF-QNRF dataset, it reduces the MAE and MSE values of the best method (\textsc{CL-CNN}) by 43.3 and 36.2, respectively. It  is  worth  mentioning that our method does not use any external detection models or multi-scale structures.
		\item \textsc{Bayesian+} consistently improves the performance of \textsc{Bayesian} by around 3\% on all the four datasets.
		\item Both \textsc{Bayesian} and \textsc{Bayesian+} outperform \textsc{Baseline} significantly on all the four datasets. \textsc{Bayesian+} makes 15\% improvements on UCF-QNRF, 9\% on ShanghaiTechA, 8\% on ShanghaiTechB, and 8\% on UCF\_CC\_50, respectively. 
	\end{itemize} 
	
	{\flushleft \textbf{Visualization of the estimated density maps.}}
	We visualize the estimated density maps using different training losses in Fig.~\ref{fig:viz}. From the close-ups we can see that \textsc{Baseline} often predicts abnormally values in the congested areas, in contrast, our \textsc{Bayesian} and \textsc{Bayesian+} give more accurate estimations. Our methods benefit from the proposed probability model which constructs soft posterior probabilities if the pixel is close to several head points. In sparse areas, on the other hand, \textsc{Baseline} can not recognize each person well, while our methods predict more accurate results both on count estimation and localization.
	\subsection{Ablation Studies}
	
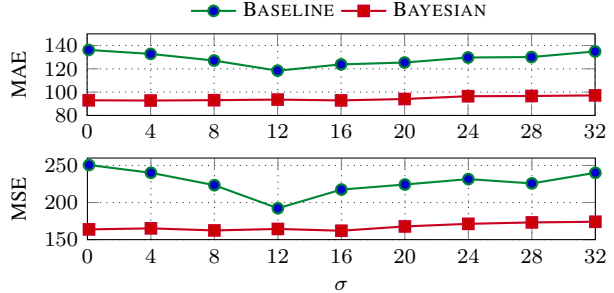
\begin{figure}[!t]
		\centering
		\begin{tikzpicture}[/pgfplots/width=1.0\linewidth,/pgfplots/height=0.32\linewidth]
		\begin{axis}
		[ymin=80,ymax=150,xmin=0,xmax=32,font=\footnotesize,
		ylabel={MAE},
		xtick={0,4,8,12,16,20,24,28,32},
		legend style={draw=none},
		legend columns=-1,
		legend entries={\textsc{Baseline},\textsc{Bayesian}},
		legend to name=legend_sigma,
		title={\ref{legend_sigma}},
		title style={yshift=-1.8ex},
		grid=both,
		grid style=dotted,
		major grid style={white!20!black},
		minor grid style={white!70!black}]
		\addplot+[mygreen,thick] coordinates {(0.1,136.20) (4,132.80) (8,127.10) (12,118.40) (16,123.80) (20,125.40) (24,129.70) (28,130.10) (32,134.90)};
		\label{curve:baseline_sigma}
		\addplot+[myred,thick]  coordinates {(0.1,93.00) (4,92.80) (8,93.10) (12,93.60) (16,92.90) (20,94.10) (24,96.50) (28,96.70) (32,97.20)};
		\label{curve:bayesian_sigma}
		\end{axis}
		\end{tikzpicture} \\ 
		\begin{tikzpicture}[/pgfplots/width=1.0\linewidth,/pgfplots/height=0.32\linewidth]
		\begin{axis}
		[ymin=150,ymax=260,xmin=0,xmax=32,font=\footnotesize,
		xlabel={$\sigma$},
		ylabel={MSE},
		xtick={0,4,8,12,16,20,24,28,32},
		legend style={at={(0.03,0.5)},anchor=west},
		legend cell align=left,
		grid=both,
		grid style=dotted,
		major grid style={white!20!black},
		minor grid style={white!70!black}]
		\addplot+[mygreen,thick] coordinates {(0.1,250.60) (4,240.10) (8,223.40) (12,192.30) (16,217.50) (20,224.30) (24,231.60) (28,225.80) (32,240.10)};
		\addplot+[myred,thick]  coordinates {(0.1,163.80) (4,165.20) (8,162.40) (12,164.40) (16,162.00) (20,167.80) (24,171.30) (28,173.20) (32,174.10)};
		\end{axis}
		\end{tikzpicture}\\
		\caption{\small{\emph{The curves of testing results for different losses \wrt the Gaussian parameter $\sigma$ on UCF-QNRF. }}}
		\label{fig:sigma}
	\end{figure}

	{\flushleft \textbf{Effect of $\sigma$.}}
	Both the proposed \textsc{Bayesian} and the \textsc{Baseline} methods  use the parameter $\sigma$ for the Gaussian kernel.
	In our loss function, $\sigma$ controls the softness of the posterior label probability as shown in Fig.~\ref{fig:entropy}, while it determines the crowd density distribution directly in \textsc{Baseline}. 
	In this subsection, we study the effect of $\sigma$ on the two methods by computing their MAE and MSE values  \wrt different $\sigma$ values on UCF-QNRF. As can be seen from the curves in Fig.~\ref{fig:sigma}:
	\begin{itemize}
		\setlength\itemsep{0em}
		\item Our \textsc{Bayesian} performs well in a wide range of values of $\sigma$. Our MAE and MSE is less than 98.0 and 180.0 when $\sigma$ changes from 0.1 to 32.0.
		\item \textsc{Baseline} is more sensitive to this parameter and its MAE and MSE vary from 118.4 to 136.2 and from 192.3 to 250.6, respectively.
	\end{itemize}
	
	\hypersetup{linkcolor=black}
	\hypersetup{linkcolor=red}
	
\begin{figure}[!h]
		\footnotesize
		\begin{tikzpicture}[/pgfplots/width=1.0\linewidth,/pgfplots/height=0.32\linewidth]
		\begin{axis}
		[ymin=80,ymax=110,xmin=0.02,xmax=1,font=\footnotesize,
		xmode = log, log basis x={e}, 
		ylabel={MAE},
		xtick={0.02,0.03,0.04,0.05,0.06,0.07,0.08,0.09,0.1,0.2,0.3,0.4,0.5,0.6,0.7,0.8,0.9,1.0},
		xticklabels={2\%,3\%,,,,,,,10\%,20\%,,,,,,,,100\%},
		legend style={cells={align=center}},
		legend style={draw=none},
		legend columns=-1,
		legend entries={\textsc{Baseline},\textsc{Bayesian},\textsc{Bayesian+}},
		legend to name=legend_margin,
		title={\ref{legend_margin}},
		title style={yshift=-2.0ex},
		grid=both,
		grid style=dotted,
		major grid style={white!20!black},
		minor grid style={white!70!black}]
		\addplot[mygreen,thick] coordinates{(0.02,106.8) (1.0,106.8)};
		\addplot[myred,thick] coordinates{(0.02,92.9) (1.0,92.9)};
		\addplot+[myblue,thick] coordinates {(0.02,105.8) (0.025,94.5) (0.03,89.4) (0.04,91.9) (0.05,86.6) (0.07,87.7) (0.10,89.5) (0.15,88.7) (0.20,88.4) (0.30,87.0) (0.60,90.1) (1.0,92.53)};
		\end{axis}
		\end{tikzpicture} \\
		\begin{tikzpicture}[/pgfplots/width=1.0\linewidth,/pgfplots/height=0.32\linewidth]
		\begin{axis}
		[ymin=135,ymax=190,xmin=0.02,xmax=1,font=\footnotesize,
		xmode = log, log basis x={e}, 
		xtick={0.02,0.03,0.04,0.05,0.06,0.07,0.08,0.09,0.1,0.2,0.3,0.4,0.5,0.6,0.7,0.8,0.9,1.0},
		xticklabels={2\%,3\%,,,,,,,10\%,20\%,,,,,,,,100\%},
		xlabel={Percentage of Image Size (x axis is in log mode)},
		xlabel shift={-1mm},
		ylabel={MSE},
		grid=both,
		grid style=dotted,
		major grid style={white!20!black},
		minor grid style={white!70!black}]
		\addplot[mygreen,thick] coordinates{(0.02,183.7) (1.0,183.7)};
		\addplot[myred,thick] coordinates{(0.02,163.0) (1.0,163.0)};
		\addplot+[myblue,thick] coordinates {(0.02,174.8) (0.025,165.7) (0.03,160.1) (0.04,162.7) (0.05,152.3) (0.07,153.7) (0.10,155.3) (0.15,154.8) (0.20,153.6) (0.30,155.7) (0.60,155.9) (1.0,162.1)};
		\end{axis}
		\end{tikzpicture}
		\caption{\small{\emph{The performance of our \textsc{Bayesian+} method \wrt $d$.}}}\label{fig:margin}
	\end{figure}
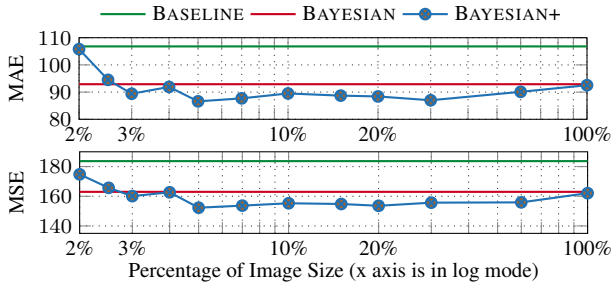

	{\flushleft \textbf{Effect of $d$.}}
	The proposed \textsc{Bayesian+} method introduces an additational parameter $d$ to control the margin between foreground and background. Fig.~\ref{fig:margin}~shows the performance of \textsc{Bayesian+} \wrt $d$ where we can conclude that: 
	\begin{itemize}
		\setlength\itemsep{0em}
		\item Our \textsc{Bayesian+} method performs well in a wide range of of values of $d$. \textsc{Bayesian+} consistently outperforms \textsc{Bayesian} when $d$ is from $3\%$ to $100\%$ of the shorter side of the image.
		\item The parameter $d$ has the meaning that the had size would not exceed $d$ so that $d$ should not be too small.
	\end{itemize}
	
	\hypersetup{linkcolor=black}
	\hypersetup{linkcolor=red}
	
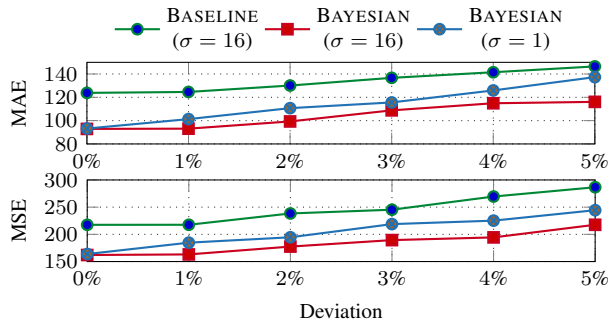
\begin{figure}[!h]
		\centering
		\begin{tikzpicture}[/pgfplots/width=1.0\linewidth,/pgfplots/height=0.32\linewidth]
		\begin{axis}
		[ymin=80,ymax=150,xmin=0,xmax=5,font=\footnotesize,
		ylabel={MAE},
		xticklabel={\pgfmathprintnumber\tick\%},
		legend style={cells={align=center}},
		legend style={draw=none},
		legend columns=-1,
		legend entries={\textsc{Baseline}\\($\sigma=16$),\textsc{Bayesian}\\($\sigma=16$),\textsc{Bayesian}\\($\sigma=1$)},
		legend to name=legend_robust,
		title={\ref{legend_robust}},
		title style={yshift=-2.0ex},
		grid=both,
		grid style=dotted,
		major grid style={white!20!black},
		minor grid style={white!70!black}]
		\addplot+[mygreen,thick]    coordinates {(0,123.8) (1,124.6) (2,130.1) (3,136.7) (4,141.5) (5,146.5)};
		\label{curve:baseline_robust}
		\addplot+[myred,thick]     coordinates {(0,92.9)  (1,93.1) (2,99.3) (3,108.8) (4,114.9) (5,116.1)};
		\label{curve:bayesian_robust_16}
		\addplot+[myblue,thick] coordinates {(0,93.2)  (1,101.3)  (2,110.7) (3,115.6) (4,125.9) (5,137.3)};
		\label{curve:bayesian_robust_1}
		\end{axis}
		\end{tikzpicture}\\
		\vspace{-2mm}
		\begin{tikzpicture}[/pgfplots/width=1.0\linewidth,/pgfplots/height=0.32\linewidth]
		\begin{axis}
		[ymin=150,ymax=300,xmin=0,xmax=5,font=\footnotesize,
		xlabel={Deviation},
		ylabel={MSE},
		xticklabel={\pgfmathprintnumber\tick\%},
		grid=both,
		grid style=dotted,
		major grid style={white!20!black},
		minor grid style={white!70!black}]
		\addplot+[mygreen,thick]    coordinates {(0,217.5) (1,217.6) (2,238.4) (3,245.3) (4,269.5) (5,286.6)};
		\addplot+[myred,thick]     coordinates {(0,162.0) (1,163.1) (2,177.5) (3,189.4) (4,194.4) (5,217.6)};
		\addplot+[myblue,thick] coordinates {(0,163.8) (1,184.7) (2,194.6) (3,218.8) (4,225.2) (5,244.4)};
		\end{axis}
		\end{tikzpicture}
		\caption{\small{\emph{Robustness evaluations to annotation error. We simulate human labeling errors by adding uniform random noises to the annotated point locations (percentage of image height).}}}
		\label{fig:robustness}
	\end{figure}

	{\flushleft \textbf{Robustness to annotation error.}} 
	In this subsection, we discuss the robustness of different loss functions \wrt annotation error. Labeling a person by a single point is ambiguous, because the person occupies an area in the image. Although most of the datasets place the annotation point at the center of each head, small errors from human labeling is inevitable.
	In this experiment, we simulate human labeling errors by adding uniform random noises to the original head positions, and test the performance of different losses at several noise levels. Since $\sigma$ is the spatial variance of Gaussian distribution, a larger $\sigma$ is helpful to tolerate such spatial noises. Therefore, we evaluate our \textsc{Bayesian} with $\sigma=1$ and $\sigma=16$, respectively, and \textsc{Baseline} with $\sigma=16$.
	As can be seen from Fig.~\ref{fig:robustness}, the proposed \textsc{Bayesian} performs better than \textsc{baseline} in different noise levels, even with a smaller $\sigma$ value.
	
	\hypersetup{linkcolor=black}
	\hypersetup{linkcolor=red}
	
	{\flushleft \textbf{Cross-dataset evaluation.}} 
	To further explore the generalization ability of different loss functions, we conduct cross-dataset experiments with the VGG-19 network. In this experiment, models are trained on one dataset and tested on the others without any further fine-tuning. More specifically, we train models on the largest UCF-QNRF dataset, and test them on UCF\_CC\_50, ShanghaiTechA and ShanghaiTechB, respectively. As can be seen from Table.~\ref{tab:transfer}, our methods have certain generalization ability and outperform \textsc{Baseline} on all datasets.
	
	\def\arraystretch{0.6}
	\begin{table}[!h]
		\begin{center}
			\small
			\setlength{\tabcolsep}{2.5pt}
			\begin{tabular}{@{}l|cc|cc|cc@{}}
				\toprule[1.0pt]
				UCF-QNRF~$\rightarrow$ &  \multicolumn{2}{c|}{ShanghaiTechA} & \multicolumn{2}{c|}{ShanghaiTechB} &
				\multicolumn{2}{c}{UCF\_CC\_50} \\
				Methods & MAE & MSE & MAE & MSE & MAE & MSE \\
				\midrule[0.5pt]
				\textsc{Baseline}  & 73.4 & 136.3 & 18.5 & 30.9 & 323.9 & 558.0 \\
				Our \textsc{Bayesian}  & 71.7 & 124.3 & 16.3 & 27.8 & 312.6 & 540.3 \\
				Our \textsc{Bayesian+} & \textbf{69.8} & \textbf{123.8} & \textbf{15.3} & \textbf{26.5} & \textbf{309.6} & \textbf{537.1} \\
				\bottomrule[1.0pt]
			\end{tabular}
			\vspace{2mm}
			\caption{\small{\emph{Quantitative results for cross-dataset evaluation. The models are trained on UCF-QNRF while tested on other datasets.}}}\label{tab:transfer}
		\end{center}
	\end{table}
	
	{\flushleft \textbf{Limiting the image resolution.}}
	We have found that image resolutions of UCF-QNRF vary widely, and a single CNN model can not handle such a large variation well. Therefore, we limit the image resolution to 2048 pixels and this experiment performs ablation study on this factor. As can be seen from Table.~\ref{tab:resize}, all methods benefit from resizing and our methods outperform the baseline method in both settings.
	
	\def\arraystretch{0.6}
	\begin{table}[!h]
		\centering
		\small
		\setlength{\tabcolsep}{11pt}
		\begin{tabular}{@{}l|cc|cc@{}}
			\toprule[1.0pt]
			~&\multicolumn{2}{c|}{With Resize} &\multicolumn{2}{c}{Without Resize} \\
			Methods & MAE & MSE & MAE & MSE  \\
			\midrule[0.5pt]
			\textsc{Baseline} & 106.8 & 183.7 & 128.7 & 193.8  \\
			Our \textsc{Bayesian} & 92.9 & 163.0 & 115.1 & 187.0 \\
			Our \textsc{Bayesian+} & \textbf{88.7} & \textbf{154.8} & \textbf{112.6} & \textbf{181.1} \\
			\bottomrule[1.0pt]
		\end{tabular}
		\vspace{2mm}
		\caption{\small{\emph{The effect of limiting image resolution on UCF-QNRF.}}}
		\label{tab:resize}
	\end{table}
	
	{\flushleft \textbf{Different backbones.}}
	Our proposed loss functions can be readily applied to any network structure to improve its performance on the crowd counting task. Here we apply the proposed losses to both VGG-19 and AlexNet and make comparisons with the baseline loss. The quantitative results from Table~\ref{tab:backbone} indicate that our Bayesian loss functions outperform the baseline loss significantly on both the networks.
	
	\def\arraystretch{0.6}
	\begin{table}[!h]
		\begin{center}
			\small
			\setlength{\tabcolsep}{11.5pt}
			\begin{tabular}{@{}l|cc|cc@{}}
				\toprule[1.0pt]
				Backbones &\multicolumn{2}{c|}{VGG-19~\cite{DBLP:journals/corr/SimonyanZ14a}} &\multicolumn{2}{c}{AlexNet~\cite{NIPS2012_4824}}  \\
				Methods & MAE & MSE & MAE & MSE  \\
				\midrule[0.5pt]
				\textsc{Baseline} & 106.8 & 183.7 & 130.8 & 221.0\\
				Our \textsc{Bayesian} & 92.9 & 163.0 & 121.2 & 202.5 \\
				Our \textsc{Bayesian+} & \textbf{88.7} & \textbf{154.8} & \textbf{116.3} & \textbf{191.7} \\
				\bottomrule[1.0pt]
			\end{tabular}
			\vspace{2mm}
			\caption{\small{\emph{Performances of models using different backbones on UCF-QNRF. Our proposed methods outperform the baseline method consistently.}}}\label{tab:backbone}
		\end{center}
	\end{table}
	
	\vspace{-2mm}
	\section{Conclusions and Future Work}
	
	In this paper, we propose a novel loss function for crowd count estimation with point supervision. Different from previous methods that transform point annotations into the ``ground-truth" density maps using the Gaussian kernel with pixel-wise supervision, our loss function adopts a more reliable supervision on the count expectation at each annotated point. 
	Extensive experiments have demonstrated the advantages of our proposed methods in terms of accuracy, robustness and generalization. The current form of our formulation is fairly general and can easily incorporate other knowledge, \eg, specific foreground or background priors, scale and temporal likelihoods, and other facts to further improve the proposed method. 
	
	{\flushleft \textbf{Acknowledgements.}}
	This work was supported by the National Basic Research Program of China (Grant No. 2015CB351705) and National Major Project of China (Grant No. 2017YFC0803905).

	{\small
		\bibliographystyle{ieee_fullname}
		\bibliography{egbib}
	}
	
\end{document}